\documentclass[10pt,twocolumn,letterpaper]{article}

\usepackage{cvpr}
\usepackage{times}
\usepackage{epsfig}

\usepackage{graphicx}
\usepackage{amsmath}
\usepackage{amssymb}
\usepackage{booktabs}
\usepackage{multirow}
\usepackage{float}
\usepackage{tabularx}
\usepackage{pifont}
\usepackage[most]{tcolorbox}
\usepackage{subfig}

\usepackage{tikz}
\newcommand*\circled[1]{\tikz[baseline=(char.base)]{\node[shape=circle,draw,inner sep=0.6pt] (char) {#1};}}

\usepackage[pagebackref,breaklinks,colorlinks]{hyperref}

\usepackage[capitalize]{cleveref}
\crefname{section}{Sec.}{Secs.}
\Crefname{section}{Section}{Sections}
\Crefname{table}{Table}{Tables}
\crefname{table}{Tab.}{Tabs.}

\cvprfinalcopy 

\def\cvprPaperID{1} 

\begin{document}

\title{Goal-driven Self-Attentive Recurrent Networks for Trajectory Prediction}

\author{Luigi Filippo Chiara$^1$\hskip 1em Pasquale Coscia$^1$\hskip 1em Sourav Das$^1$\hskip 1em Simone Calderara$^2$\\Rita Cucchiara$^2$\hskip 1em  Lamberto Ballan$^1$\\\\
$^1$University of Padova, Italy\hskip 1em $^2$University of Modena and Reggio Emilia, Italy\\
{\tt\small \{luigifilippo.chiara, pasquale.coscia\}@unipd.it \quad sourav.das@phd.unipd.it}\\
{\tt\small \{simone.calderara, rita.cucchiara\}@unimore.it \quad lamberto.ballan@unipd.it}
}


\maketitle


\begin{abstract}
Human trajectory forecasting is a key component of autonomous vehicles, social-aware robots and advanced video-surveillance applications. This challenging task typically requires knowledge about past motion, the environment and likely destination areas. In this context, multi-modality is a fundamental aspect and its effective modeling can be beneficial to any architecture. Inferring accurate trajectories is nevertheless challenging, due to the inherently uncertain nature of the future. To overcome these difficulties, recent models use different inputs and propose to model human intentions using complex fusion mechanisms. In this respect, we propose a lightweight attention-based recurrent backbone that acts solely on past observed positions. Although this backbone already provides promising results, we demonstrate that its prediction accuracy can be improved considerably when combined with a scene-aware goal-estimation module. To this end, we employ a common goal module, based on a U-Net architecture, which additionally extracts semantic information to predict scene-compliant destinations. We conduct extensive experiments on publicly-available datasets (i.e. SDD, inD, ETH/UCY) and show that our approach performs on par with state-of-the-art techniques while reducing model complexity.
\end{abstract}


\section{Introduction}
\label{sec:introduction}

\begin{figure}[ht]
    \includegraphics[width=\columnwidth]{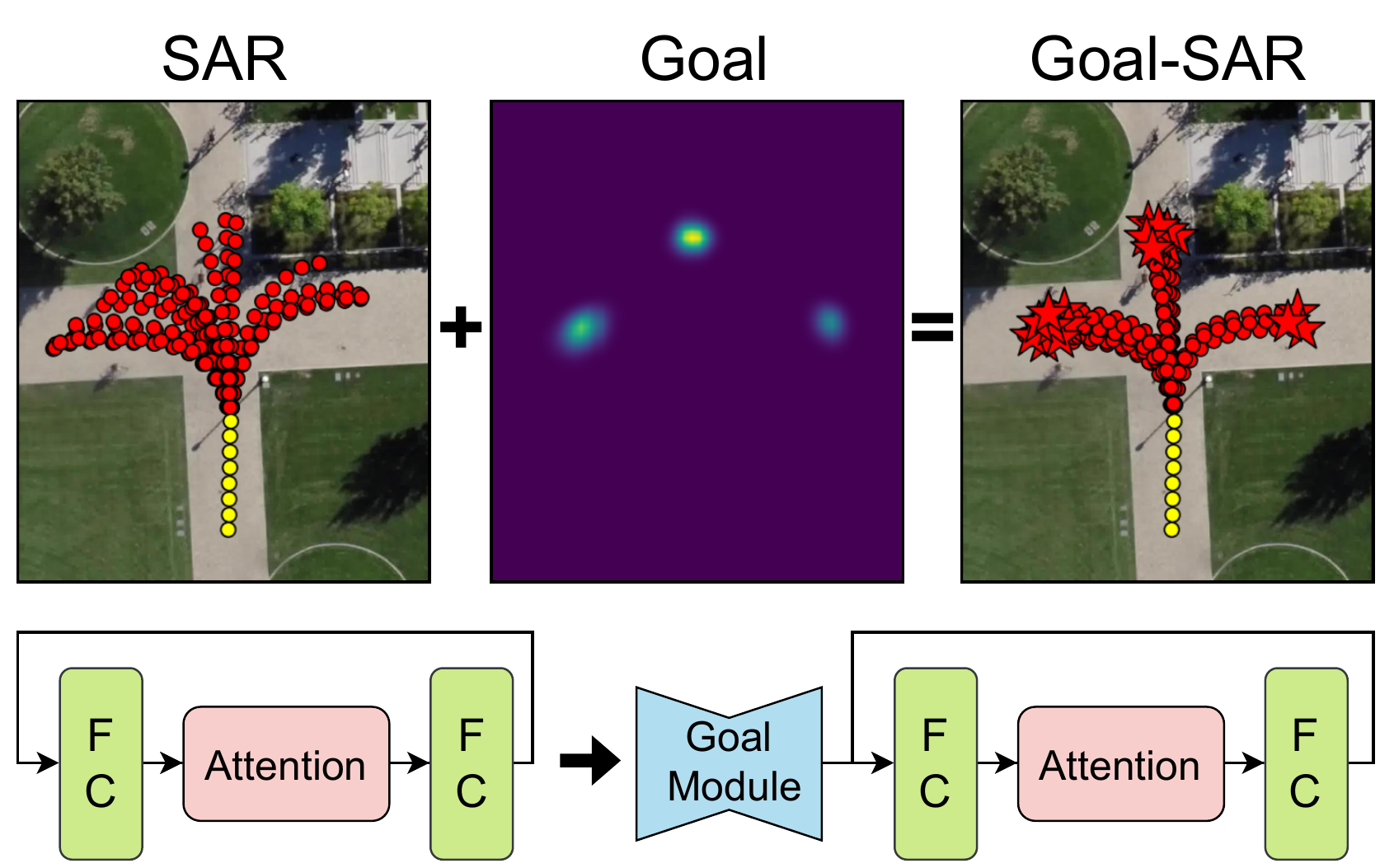}
    \caption{
    Multiple paths can occur in the future in specific circumstances (\eg people approaching roundabouts or intersections). These situations may lead a prediction model to learn an ``average" behavior envisioning not feasible trajectories. To overcome this issue, we employ a semantic-aware goal module to infer multiple likely destinations that are then fed to a temporal backbone to predict scene-complaint and multi-modal outputs.}
    \label{fig:intro}
\end{figure}

Human trajectory forecasting has aroused great interest in several scientific communities (\eg computer vision, robotics and intelligent transportation) in recent years~\cite{Kitani_2012_ECCV, rudenko_survey, Social-lstm, KTMP16, mantra} since it involves human perception, motion analysis and reasoning. Predicting human dynamics is essential for systems that need to proactively react to the surrounding environment. For example, autonomous vehicles need to foresee future positions to avoid collisions, while robots require to behave in a socially-compliant way to move naturally alongside humans. Several video surveillance tasks may also gain an advantage from motion prediction to detect anomalous behaviors and support human operators in order to intervene promptly.

Pedestrian motion prediction mainly relies on previously observed locations to infer future positions and, at its core, it boils down to the prediction of a time series. When urban areas are considered, this task can become extremely complex, due to the large number of involved factors. For instance, people may prefer a right-of-way rule in scarcely crowded areas but may not follow it when the number of agents increases. Past methods~\cite{social_force, constant_velocity} have conventionally tackled these challenges by adopting simple motion models and hand-crafted functions, which may be effective only under ideal conditions. However, the recent availability of large-scale annotated datasets~\cite{sdd, inDdataset} has shifted this modeling paradigm towards data-driven approaches~\cite{Social-lstm, social-gan, Dendorfer_2021_ICCV, Lisotto_2019_ICCV, Vemula2017SocialAM}, contributing to notable performance improvement.

While collected positions can be analyzed as temporal sequences, taking advantage of additional information can improve prediction accuracy. This auxiliary information can be extracted from several sources: scene description (\ie RGB images), presence of social interactions, and possibly also other modes (\eg semantic information, depth estimation or videos). Employing these supplementary cues, social-aware~\cite{Social-lstm, Yuan_2021_ICCV} and environment-aware~\cite{Sadeghian_2018_ECCV, Sadeghian_2019_CVPR, Choi_2019_ICCV, Liang_2020_CVPR} prediction methods have been developed. More recently, attention mechanisms have proved their effectiveness in forecasting human locations, following recent advances in natural language processing (NLP) domains~\cite{vaswani_attention, NEURIPS2019_dc6a7e65}.

Multi-modal approaches have also been investigated to capture the inherently multi-modal nature of the task~\cite{social-gan, trajectronpp}. Recent methods explicitly model this aspect in terms of \emph{goal} prediction~\cite{dendorfer2020goalgan, mangalam2020pecnet, Mangalam_2021_ICCV, Dendorfer_2021_ICCV, Zhao_2021_ICCV}. Indeed, goal-conditioned models represent a class of approaches investigated in several fields, \eg reinforcement learning~\cite{goal_rl, goal_rl2} and motion planning~\cite{DBLP:conf/icml/Kaelbling93, ziebart2009planning}. Trajectory forecasting methods should therefore: (i) include a sequential processing of observed locations, and (ii) deal with the intrinsic uncertain and multi-modal nature of the future.

To tackle the above challenges, we propose an attention-based approach conditioned on estimated destinations to infer multi-modal scene-compliant predictions. Our approach empowers a recurrent trajectory forecasting backbone with an additional goal-estimation module, which takes as inputs motion history and pixel-level semantic classes. This information is fed to a U-Net~\cite{u-net} architecture that outputs probability distribution maps from which likely future goals are then sampled. We provide these cues to the temporal backbone that, conditioned on estimated final positions, predicts subsequent time steps in a recurrent fashion. An overview of our approach is provided in Fig.~\ref{fig:intro}. Compared to prior work~\cite{dendorfer2020goalgan}, in which estimated goals represent hard constraints, we propose a soft-constrained model that automatically learns to use future goals to infer human paths.

To summarize, our contribution is threefold. \textbf{First}, we propose a simple yet effective recurrent backbone based on a multi-head attention mechanism, that is able to outperform several recent methods by solely leveraging past observed positions. \textbf{Second}, we demonstrate that these results can be considerably enhanced with the addition of a scene-aware goal-estimation module. This highlights that multi-modality is a fundamental component of the prediction task.
\textbf{Third}, we present experimental results on a variety of benchmarks (\ie Stanford Drone~\cite{sdd}, Intersection Drone~\cite{inDdataset} and ETH/UCY datasets), and provide additional insights on the accuracy of the goal-estimation module with comprehensive ablations studies. 
Code available at \href{https://github.com/luigifilippochiara/Goal-SAR}{https://github.com/luigifilippochiara/Goal-SAR}.


\section{Related Work}
\label{sec:related}

\begin{figure*}
\centering
  \includegraphics[width=0.84\textwidth]{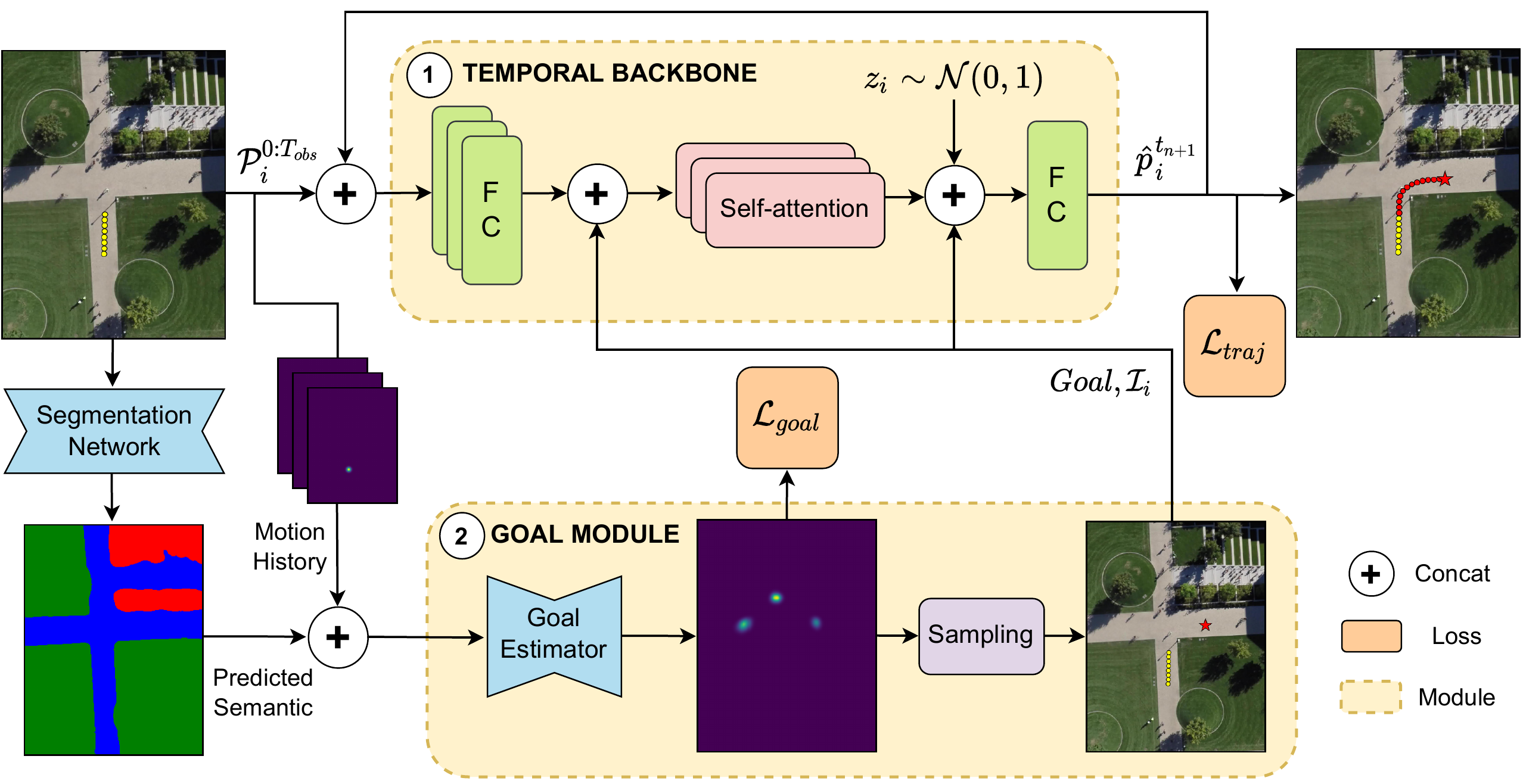}
  \caption{\textbf{Model Architecture.} Our forecasting method consists of two main components: \protect\circled{1} Temporal-Attention Backbone and \protect\circled{2} Goal Module. The recurrent backbone implements a self-attention mechanism that solely relies on previous observed positions. The Goal module processes motion history and semantic scene segmentation through a U-Net architecture to predict future goals as probability distributions. Final positions are then sampled and fed to the temporal backbone. The output, together with some random noise, is then decoded by a FC layer to extract the next predicted position at time step $t_{n+1}$.}
  \label{fig:model}
\end{figure*}

Human trajectory forecasting in crowded contexts has been extensively studied by several works. For example, Alahi \etal~\cite{Social-lstm} use a long short-term memory (LSTM) network with a social pooling layer that extracts interactions among nearby pedestrians, and Gupta \etal~\cite{social-gan} use Generative Adversarial Networks (GANs) to predict social acceptable trajectories. Other approaches~\cite{zhang2019srlstm, Mohamed_2020_CVPR, liu2020snce} propose other strategies such as state refinements, spatio-temporal graph neural networks \cite{kipf2017semi, 9046288}, and social contrastive losses.
Some scenarios may benefit from modeling, additionally, human-space interactions~\cite{Kitani_2012_ECCV, KTMP16, IVC18, Sadeghian_2019_CVPR}.
For instance, Xue \etal~\cite{ss-lstm} analyze three different scales of interaction (person, neighborhood, and scene), and Choi \etal~\cite{Choi_2019_ICCV} encodes visual spatio-temporal features where dynamics are influenced by objects and people in the scene. Lisotto \etal~\cite{Lisotto_2019_ICCV} jointly model human interactions and space perception through similar pooling mechanisms proposed in Alahi et al.~\cite{Social-lstm} based on past observations and semantic scene segmentation. Furthermore, Liang \etal~\cite{Liang_2020_CVPR} propose a multi-scale approach to firstly predict coarse and then fine-grained locations using semantic scene segmentation. Finally, Salzmann \etal~\cite{trajectronpp} enforce dynamical systems constraints into a graph-structured recurrent model designed for robotic planning and control frameworks.

\paragraph{Attention-based Models.}
Several attention mechanisms demonstrated to be more effective than traditional recurrent architectures to model temporal dependencies.
Vemula \etal~\cite{Vemula2017SocialAM} and Huang \etal~\cite{Huang_2019_ICCV} propose similar approaches based on spatial-temporal graph attention networks (GAT) where LSTM networks encode both spatial and temporal correlations in the crowd.
Sadeghian \etal~\cite{Sadeghian_2018_ECCV} propose a deep visual attention-based model which focuses on most representative areas processed by CNNs. Sadeghian \etal~\cite{Sadeghian_2019_CVPR} leverage social attention and physical attention mechanisms by fusing past motion history and scene context information.
Furthermore, Giuliari \etal~\cite{giuliari2020transformer} propose a transformer-based model predicting motion from past observed positions, while Yu \etal~\cite{YuMa2020Spatio} propose a spatio-temporal graph transformer framework that models social interactions with an encoder-decoder attention-based convolution mechanism. Yuan \etal~\cite{Yuan_2021_ICCV} propose an attentive architecture that models temporal and social dimensions preserving time and agent information.

\paragraph{Destination-based Models.}
More recently, several works leverage goal prediction to improve prediction accuracy~\cite{albrecht2021interpretable, zhao2020tnt}. Mangalam \etal~\cite{mangalam2020pecnet} employ a variational autoencoder (VAE) conditioned on estimated endpoints, where a non-local attention mechanism recursively updates hidden state representations. Similarly, Dendorfer \etal~\cite{dendorfer2020goalgan} investigate multi-modality conditioning routes to estimated final positions. The limitation of providing continuous non-zero distributions is then addressed by Dendorfer \etal~\cite{Dendorfer_2021_ICCV}, which propose to train multiple generators, each one predicting a different distribution associated with a specific mode. Similar to ours, Mangalam \etal~\cite{Mangalam_2021_ICCV} propose a convolutional-based approach where positions are treated as heat-maps and final positions are sampled from 2D probability distribution maps. Finally, Zhao and Wildes~\cite{Zhao_2021_ICCV} propose an LSTM encoder-decoder architecture where predictions are conditioned upon destinations retrieved from an expert repository.


\section{Proposed Approach}
\label{sec:method}

Pedestrian trajectory forecasting in urban scenarios is a challenging task requiring to model multiple factors that influence human motion. We argue that multi-modality modeling (here instantiated in terms of goal prediction) is a key component of this problem and can enhance almost any model. Indeed, predicting likely destinations (\ie goals) may steer paths towards more realistic locations and allow an elegant and effective factorization of the intrinsic multi-modality of the future. To this end, we conceive a simple yet powerful architecture that consists of two main components: a temporal recurrent backbone (Fig.~\ref{fig:model} - \circled{1}), which processes locations using a self-attentive mechanism, and a goal-estimation module (Fig.~\ref{fig:model} - \circled{2}), which predicts likely final locations of each agent given its previously observed positions. Our model is thus able to predict plausible trajectories in complex scenarios processing both temporal dependencies and future intentions.

\subsection{Recurrent Backbone}
\paragraph{Trajectory embedding.}
Our backbone is a recurrent architecture that only relies on temporal information. It processes sequences of 2D locations $p_i^t =(x_i^t, y_i^t)$ for the $i^{th}$ agent at time $t$. Sequences span from $t=0$ to $t=T_{obs} + T_{pred} = T_{seq}$, where $T_{obs}$ (resp. $T_{pred}$) is the observation (resp. prediction) time window, and $T_{seq}$ is the total sequence length. Every agent is considered independently from others. Input coordinates $\mathcal{P}_i^{0:t_{n}} = (p_i^{0}, ..., p_i^{t_n})$ from $t=0$ to $t_n$ are encoded into a feature space as follows:

\begin{equation}
    e_i^t = \phi(p_i^t; W_e),
\end{equation}

where $\phi(\cdot)$ is a linear embedding function with ReLU non-linearity and $W_e$ embedding weights. As in Zhang \etal~\cite{zhang2019srlstm}, we subtract the last observed position from each processed sequence for data normalization.

\paragraph{Temporal Attention.}
To process temporal dependencies, transformers~\cite{vaswani_attention} have recently aroused great success due to their speed and performance. Their encoder-decoder paradigm models relationships between every pair of input/output sequences stacking multiple identical layers, using a multi-head self-attention mechanism and fully-connected feed-forward networks. Inspired by Yu \etal~\cite{YuMa2020Spatio}, we only leverage the encoding part, with the aim to predict future positions given variable-length input sequences $\mathcal{P}_i^{0:t_{n}} = (p_i^{0}, ..., p_i^{t_n})$, in a recurrent fashion. More explicitly, temporal dependencies across subsequent time steps are taken into account by linearly projecting embedded positions $(e_i^{0}, ..., e_i^{t_n})$ into three different vectors: $q_i^t$ (query), $k_i^t$ (key) and $v_i^t$ (value). A dot product between queries and values is performed to compute attention coefficients used to weight the values $v_i^t$ and provide the corresponding output as follows:

\begin{equation}
\label{eq:attention}
    ATT(Q_i, K_i, V_i) = \textrm{softmax}\bigg(\frac{Q_iK_i^T}{\sqrt d_k}\bigg) V_i,
\end{equation}

where $d_k$ represents a normalization factor. This operation is performed $N_{head}$ times using different linear projections of $Q_i$, $K_i$ and $V_i$, yielding a vector of new embedded positions $(h_i^{0}, ..., h_i^{t_n})$, which incorporate temporal dependencies.

\paragraph{Trajectory decoder.}
The last encoded feature vector $h_i^{t_n}$ is then fed to a decoder defined by a linear layer $\psi(\cdot)$ with ReLU non-linearity and weights $W_d$, to extract decoded positions at time $t_{n+1}$. To increase the variance of generated sequences, we concatenate a Gaussian random vector $z_i \sim \mathcal{N}(0, \textbf{I}_z)$ to this hidden state as in Huang \etal~\cite{Huang_2019_ICCV}:

\begin{equation}
    \hat{p}_i^{t_{n+1}} = \psi\big(\textrm{concat}(h_i^{t_n}, z_i); W_d\big).
\end{equation} 

In the following, we refer to our Self-Attentive Recurrent backbone as SAR. It predicts the next position at $t+1$ using all previous time steps. Instead of using a hidden state to keep track of previous inputs, typically employed in recurrent architectures~\cite{zhang2019srlstm, Social-lstm}, we recursively concatenate estimated locations to the previous input sequence. Our temporal module is depicted in Fig.~\ref{fig:model} as \circled{1}.

\vspace{0.5cm}

\subsection{Goal Module}
\label{sec:goal}
The purpose of the goal module is to model the multi-modality of human motion, and this is achieved by predicting a probability distribution of plausible final positions (\ie goals) for each input trajectory.
We use the same goal module proposed in Mangalam \etal~\cite{Mangalam_2021_ICCV}, slightly modifying its pre-processing steps and output format. This module concatenates both observed positions and visual scene information, which are then fed to a U-Net~\cite{u-net} model that directly outputs a probability map of future final locations.

\paragraph{Scene semantic.} 
Scene context is an important factor to consider for estimating more realistic paths. For example, obstacles may influence human dynamics or sidewalks may be the natural choice for pedestrians rather than roads. To this end, semantic information is extracted from bird's-eye view RGB images using an off-the-shelf pre-trained semantic segmentation network taken from Mangalam \etal~\cite{Mangalam_2021_ICCV}. More specifically, the following set of semantic classes $C=\{$\texttt{pavement, terrain, structure, tree, road, not defined}$\}$ are considered, resulting in a semantic tensor $\mathcal{S} \in \mathbb{R}^{W \times H \times C}$ where $W$ and $H$ represent input image sizes. Each slice of $\mathcal{S}$ represents a specific semantic class containing 0's or 1's labeling each pixel with the corresponding class.

\paragraph{Goal Encoder-Decoder.} 
The semantic tensor $\mathcal{S}$ is concatenated to $N_{obs}$ distribution maps depicting past motion history. More precisely, for each observed position we consider a heat-map of spatial sizes $W$ and $H$, and create a 2D Gaussian probability distribution map with mean $p_i^t$ and variance $\sigma_S^2 {\bf{I}}_2$. We denote with $\mathcal{M}$ the projection from 2D $(x,y)$ coordinates to $W \times H$ heat-map representations. After concatenation, we obtain a $W \times H\times (C + N_{obs})$ trajectory-on-scene input tensor ${\bf{H}}_S$.

This tensor is then fed to a U-Net architecture consisting of $L$ blocks that reduce input spatial dimension $H\times W$ using double convolutional layers with ReLU non-linearity and max-pooling operations. Each intermediate output $\bf{H}_l$ ($1 \leq l \leq L$) is then passed via skip-connections to the decoder. In the expanding arm, $L$ decoder blocks process ${\bf{H}}_L$, doubling its resolution using bilinear up-sampling, double convolutions and ReLU non-linearity. Skip connections fuse ${\bf{H}}_l$ tensors from the contracting arm, and a final output convolutional layer followed by a pixel-wise sigmoid return the spatial probability distribution of the final position $\mathcal{M}(p^{T_{seq}})$. The goal-estimation module is depicted in Fig.~\ref{fig:model} as \circled{2}.

\paragraph{Distribution Sampling.}
The goal module outputs a 2D heat-map which represents the probability of the monitored agent to be in a specific final location at $T_{obs} + T_{pred}$ given the information from $t=0$ to $T_{obs}$. Estimated goals are sampled from these probability maps. When more than one sample is required, we find it beneficial to use the Test-Time-Sampling-Trick \texttt{TTST} proposed in Mangalam \etal~\cite{Mangalam_2021_ICCV}, where $10,000$ goals are initially sampled and then clustered with K-means to obtain the $20$ output modalities. To inject the estimated destination into our temporal backbone, we concatenate to $\mathcal{P}_i^{0:t_{n}}$ the goal and the following three additional inputs, denoted as $\mathcal{I}_i$: last position, current distance to the estimated goal and time step value $t$, as in Dendorfer \etal~\cite{dendorfer2020goalgan}. We also find it beneficial to use a skip connection to feed our backbone with this additional information, which is concatenated both before and after the self-attention layer, as shown in Fig.~\ref{fig:model}.
This choice will be better motivated in one of the ablation studies of Section~\ref{sec:experiments}.

\subsection{Loss Function}
We consider two losses to train our architecture:

\vspace{-0.4cm}

\begin{equation}
    \mathcal{L}_{goal} = \frac{1}{N_p}\sum_{i=1}^{N_p} \textrm{BCE}\Big(\mathcal{M}(p_i^{T_{seq}}), \hat{\mathcal{M}}(p_i^{T_{seq}})\Big),
\end{equation}

\vspace{-0.5cm}

\begin{equation}
    \mathcal{L}_{traj} = \frac{1}{N_p \cdot T_{pred}}\sum_{i=1}^{N_p}\sum_{t=T_{obs} + 1}^{T_{seq}}{\left\Vert p_i^t - \hat{p}_i^t\right\Vert}_2^2.
\end{equation}

We firstly train the goal module to minimize a Binary Cross-Entropy loss between predicted and ground-truth probability maps obtained as 2D Gaussian distributions $\mathcal{N}\big(p_i^{T_{seq}}, \sigma_S^2{\bf{I}}_2\big)$ centered at ground-truth final destinations. Secondly, we train our recurrent backbone minimizing the Mean Squared Error between predicted and ground-truth positions from $T_{obs} + 1$ to $T_{seq}$. Both terms are then normalized with respect to the number of processed agents $N_p$. Our total loss function is defined as:

\begin{equation}
    \mathcal{L} = \mathcal{L}_{goal} + \lambda \mathcal{L}_{traj},
    \label{eq:loss}
\end{equation}

\noindent where $\lambda$ is a hyper-parameter balancing each network's contribution.


\section{Results}
\label{sec:experiments}

\subsection{Datasets and Experiments}
\paragraph{Experimental Settings.} We follow the well-established experimental protocol used in human trajectory prediction~\cite{Social-lstm, social-gan}, that is to observe $3.2~s$ and to predict the next $4.8~s$. This leads to consider $T_{obs} = 8$ and $T_{pred} = 12$ time steps, respectively. In the following, we compare two main models: SAR and Goal-SAR. SAR only uses the temporal attention module, while Goal-SAR also includes the goal module and its estimated final positions.

\paragraph{Datasets.}
\label{sec:datasets}
We evaluate our model on three standard pedestrian datasets: Stanford Drone Dataset (SDD)~\cite{sdd}, Intersection Drone Dataset (inD)~\cite{inDdataset}, and ETH/UCY~\cite{lerner, pellegrini}. Stanford Drone Dataset represents the first large-scale dataset proposed for human trajectory prediction, and captures several large areas of a university campus from a bird's-eye-view perspective. It is split into $60$ recordings where complex human dynamics show strong interactions with the surrounding environment. We use the same split proposed in Kothari \etal.~\cite{trajnetpp} and used by most recent works~\cite{mangalam2020pecnet, Sadeghian_2019_CVPR}, where 30 scenes are used as train and $17$ as test data, and only \textit{pedestrian} are retained. Data is down-sampled at $2.5$ FPS. inD contains $32$ recordings of trajectories collected at $4$ German intersections, where pedestrians interact with cars to reach their destinations. We filter out all non-pedestrian trajectories and consider the evaluation protocol proposed in Bertugli \etal~\cite{Bertugli2021-acvrnn}, where all scenes are split into train, validation, and test sets with a $70$-$10$-$20$ rule. Finally, ETH/UCY are the classic benchmarks for pedestrian trajectory prediction. They are sampled at $2.5$ FPS and contain five different scenes (ETH, HOTEL, UNIV, ZARA1 and ZARA2) monitoring entrances of buildings and sidewalks from RGB cameras typically used in video surveillance applications. $1536$ pedestrians are captured, mainly showing human-human interactions. We follow the common leave-one-scene-out protocol~\cite{social-gan}, using $4$ scenes for training and the remaining one for testing.

\begin{table}[t]
\centering
\resizebox{0.82\columnwidth}{!}{%
\begin{tabular}{c|ccc|cc}
    \toprule
    \textbf{Method} & \textbf{S} & \textbf{I} & \textbf{G} & \textbf{ADE} & \textbf{FDE}\\
    \cmidrule{1-6}
    Social-LSTM~\cite{Social-lstm} & \ding{51} & & & 57.00 & 31.20\\
    Social-GAN~\cite{social-gan} & \ding{51} & & & 27.23 & 41.44\\
    Goal-GAN~\cite{dendorfer2020goalgan} & & \ding{51} & \ding{51} & 12.20 & 22.10\\
    PECNet~\cite{mangalam2020pecnet} & \ding{51} & & \ding{51} & 9.96 & 15.88\\
    MG-GAN~\cite{Dendorfer_2021_ICCV} & \ding{51} & & \ding{51} & 13.60 & 25.80\\
    \textsf{Y}-net~\cite{Mangalam_2021_ICCV} & & \ding{51} & \ding{51} & \underline{7.85} & \underline{11.85}\\
    \cmidrule{1-6}
    \textbf{SAR (Ours)} & & & & 10.73 & 18.66\\
    \textbf{Goal-SAR (Ours)} & & & \ding{51} & \textbf{7.75} & \textbf{11.83}\\
    \bottomrule
\end{tabular}
}
\caption{\textbf{Stanford Drone dataset (SDD) results.} Results are reported as the minimum ADE and FDE of 20 predicted samples, in pixels (lower is better). Bold and underlined numbers indicate best and second-best. \textbf{S}, \textbf{I}, and \textbf{G} indicate additional input information (other than temporal) used by the models and represent Social component, Image and Goal-estimation module, respectively.}
\label{tab:sdd_short}
\end{table}

\begin{table}[t]
\centering
\resizebox{0.82\columnwidth}{!}{%
\begin{tabular}{c|ccc|cc}
    \toprule
    \textbf{Method} & \textbf{S} & \textbf{I} & \textbf{G} & \textbf{ADE} & \textbf{FDE}\\
    \cmidrule{1-6}
    Social-GAN~\cite{social-gan} &  \ding{51} & & & 0.48 & 0.99\\
    ST-GAT~\cite{Huang_2019_ICCV} & \ding{51} & & & 0.48 & 1.00\\
    AC-VRNN~\cite{Bertugli2021-acvrnn} & \ding{51} & & & 0.42 & 0.80\\
    \textsf{Y}-net~\cite{Mangalam_2021_ICCV}* & & \ding{51} & \ding{51} & \underline{0.34} & \underline{0.56}\\
    \cmidrule{1-6}
    \textbf{SAR (Ours)} & & & & 0.39 & 0.80\\
    \textbf{Goal-SAR (Ours)} & & & \ding{51} & \textbf{0.31} & \textbf{0.54}\\
    \bottomrule
\end{tabular}
}
\caption{\textbf{Intersection Drone dataset (inD) results.} We report results considering the minimum ADE and FDE of 20 predicted samples, in meters. Lower is better. Bold and underlined numbers indicate best and second-best. Note: * means the model was trained by us, since the authors did not perform this experiment.}
\label{tab:short_term_ind}
\end{table}

\begin{table*}[th]
\centering
\resizebox{0.82\textwidth}{!}{%
\begin{tabular}{c|ccc|ccccc|c}
    \toprule
    \multicolumn{10}{c}{Evaluation Metrics (ADE~$\downarrow$~/~FDE~$\downarrow$) on Min$_{20}$}\\
    \toprule
    \textbf{Method} & \textbf{S} & \textbf{I} & \textbf{G} & \textbf{ETH} & \textbf{HOTEL} & \textbf{UNIV} & \textbf{ZARA1} & \textbf{ZARA2} & \textbf{AVG}\\   
    \cmidrule{1-10}
    Social-LSTM~\cite{Social-lstm} & \ding{51} & & & 1.09/2.35 & 0.79/1.76 & 0.67/1.40 & 0.47/1.00 & 0.56/1.17 & 0.72/1.54\\
    Social-GAN~\cite{social-gan} & \ding{51} & & & 0.81/1.52 & 0.72/1.61 & 0.60/1.26 & 0.34/0.69 & 0.42/0.84 & 0.58/1.18\\
    ST-GAT~\cite{Huang_2019_ICCV} & \ding{51} &  &  & 0.65/1.12 & 0.35/0.66 & 0.52/1.10 & 0.34/0.69 & 0.29/0.60 &  0.43/0.83\\
    Transformer-TF~\cite{giuliari2020transformer} & & & & 0.61/1.12 & 0.18/0.30 & 0.35/0.65 & 0.22/0.38 & 0.17/0.32 &  0.31/0.55\\
    STAR~\cite{YuMa2020Spatio} & \ding{51} & & & 0.36/0.65 & 0.17/0.36 & 0.31/0.62 & 0.26/0.55 & 0.22/0.46 &  0.26/0.53\\
    Trajectron++~\cite{trajectronpp} & \ding{51} & & & 0.39/0.83 & 0.12/0.21 &\textbf{ 0.20}/0.44 & \textbf{0.15}/0.33 & \textbf{0.11}/0.25 &  \underline{0.19}/0.41\\
    AgentFormer~\cite{Yuan_2021_ICCV} & \ding{51} & & & \textbf{0.26}/\underline{0.39} & \underline{0.11}/\textbf{0.14} & 0.26/0.46 & \textbf{0.15}/\textbf{0.23} & 0.14/\underline{0.24} &  \textbf{0.18}/\underline{0.29}\\
    Goal-GAN~\cite{dendorfer2020goalgan} & & \ding{51} & \ding{51} & 0.59/1.18 & 0.19/0.35 &  0.60/1.19 & 0.43/0.87 & 0.32/0.65 & 0.43/0.85\\
    PECNet~\cite{mangalam2020pecnet} & \ding{51} & & \ding{51} & 0.54/0.87 & 0.18/0.24 & 0.35/0.60 & 0.22/0.39 & 0.17/0.30 & 0.29/0.48\\
    MG-GAN~\cite{Dendorfer_2021_ICCV} & \ding{51} & & \ding{51} & 0.47/0.91 & 0.14/0.24 & 0.54/1.07 & 0.36/0.73 & 0.29/0.60 &  0.36/0.71\\
    \textsf{Y}-net~\cite{Mangalam_2021_ICCV} & & \ding{51} & \ding{51} & \underline{0.28}/\textbf{0.33} & \textbf{0.10}/\textbf{0.14} & \underline{0.24}/\textbf{0.41} & \underline{0.17}/0.27 & \underline{0.13}/\textbf{0.22} &  \textbf{0.18}/\textbf{0.27}\\
    \cmidrule{1-10}
    \textbf{SAR (Ours)} & & & & 0.34/0.64 & 0.14/0.29 & 0.33/0.66 & 0.25/0.51 & 0.21/0.43 &  0.25/0.51\\
    \textbf{Goal-SAR (Ours)} & & & \ding{51} & \underline{0.28}/\underline{0.39} & 0.12/\underline{0.17} & 0.25/\underline{0.43} & \underline{0.17}/\underline{0.26} &  0.15/\textbf{0.22} &  \underline{0.19}/\underline{0.29}\\
    \bottomrule
\end{tabular}
}
\caption{\textbf{ETH/UCY results.} We report results considering the minimum ADE/FDE of 20 predicted samples, denoted as Min$_{20}$, in meters. Lower is better. Bold and underlined numbers indicate best and second-best. \textbf{S}, \textbf{I}, and \textbf{G} indicate additional input information (other than temporal) used by the models and represent Social component, Image and Goal-estimation module, respectively.}
\label{tab:eth_ucy}
\end{table*}

\paragraph{Metrics.} 
To quantitatively evaluate our model, we consider two standard error metrics, namely the Average Displacement Error (ADE) and the Final Displacement Error (FDE). The former measures the average $l_2$ distance between predicted and ground truth trajectories, while the latter only considers the final positions. Specifically, since we frame our analysis in a stochastic setting, we report Min$_{20}$ADE and Min$_{20}$FDE metrics, which are obtained by generating 20 predicted samples for each input trajectory and retaining the one that provides the smallest errors.

\paragraph{Implementation details.} 
We consider the same pre-processing as in Mangalam \etal~\cite{Mangalam_2021_ICCV} for all datasets. For our recurrent backbone, due to the relatively small complexity of the problem, we use an embedding dimension of size $32$ for spatial coordinates, one transformer encoder layer with 8 multi-head attention heads, and a concatenated noise $z_i$ of size $16$. Furthermore, for the goal module we use $L=5$ down- and up-sampling blocks with number of channels (32, 32, 64, 64, 64) and (64, 64, 64, 32, 32) for the encoder and decoder, respectively, as in the standard implementation~\cite{Mangalam_2021_ICCV}. To lighten the computation burden of the goal module, we down-sample the input ${\bf{H}}_S$ tensor and output probability map by a factor of 4 for ETH/UCY/inD and 8 for SDD. We set $\sigma_S = \textrm{min}(H, W)$, and we compute the Binary Cross-Entropy loss with the log-sum-exp trick~\cite{log_sum_trick}.
Moreover, we consider a batch size of $32$, set $\lambda$ to $10^{-6}$, and jointly train all modules end-to-end for $500$ epochs using Adam optimizer with a learning rate of $10^{-4}$.

To train the semantic segmentation network, we only use the corresponding images from the trajectory train scenes defined in~\ref{sec:datasets} and its evaluation is performed on the unseen test set images.
All the experiments are run on a single 16GB GPU with PyTorch implementation. A strong data augmentation pipeline composed of random rotations, $x$ and $y$ flips, small translations, shears, and perspective modifications is used to increase the number of samples for both our temporal and goal modules. This allows our model to avoid over-fitting and provide the best possible generalization capabilities. To speed up training, the recurrent part of the backbone is trained with teacher forcing. Furthermore, to force our backbone to follow the goal module predictions, we feed ground-truth goals at training time, thus effectively decoupling the two architectures.

\subsection{Quantitative Results}

\begin{table*}
\centering
\resizebox{0.9\textwidth}{!}{%
\begin{tabular}{c|ccc|ccc|ccc}
    \toprule 
    & \multicolumn{6}{c}{\textbf{(a) Backbone choice}} \vrule & \multicolumn{3}{c}{\textbf{(b) Fusion mechanism} (Goal-SAR)}\\
    \midrule
     & RNN & LSTM & SAR & 
    Goal-RNN & Goal-LSTM & Goal-SAR & Late fusion & Early fusion & Skip-connection\\
    \midrule
    \textbf{ADE} & 11.59 & 11.08 & \textbf{10.73} & 8.68 & 8.43 & \textbf{7.75} & 9.94 & 8.27 & \textbf{7.75}\\ 
    \textbf{FDE} & 20.31 & 19.96 & \textbf{18.66} & 13.93 & 13.42 & \textbf{11.83} & 17.42 & 12.33 & \textbf{11.83}\\ 
   \bottomrule
\end{tabular}
}
\caption{\textbf{Ablation study.} We report Min$_{20}$ADE and Min$_{20}$FDE metrics obtained with (a) different recurrent backbones and (b) different fusion mechanisms for the goal and additional information. Results (in pixels) are computed on SDD.\vspace{-5pt}}
\label{tab:ablation}
\end{table*}

\paragraph{SDD dataset.}
Table~\ref{tab:sdd_short} shows our results on Stanford Drone dataset. Our temporal backbone (SAR) performs on par with several recent approaches. In particular, we note that it reaches similar results to PECNet and outperforms both Goal-GAN and MG-GAN, all of which rely on a goal estimation module and use more complex spatio-temporal architectures. With the addition of the goal module, Goal-SAR is able to slightly improve \textsf{Y}-net results, even though it uses a relatively simpler architecture. We attribute these improvements to the effectiveness of our backbone, which is able to generate smoother and more flexible predictions with respect to a convolutional one. Note that in the tables we do not report Expert~\cite{Zhao_2021_ICCV} results, as it relies on goals selected to be as close as possible to ground-truth destinations, so that it is not comparable. Besides, we also report additional inputs used by the models, which we divide into \textbf{S}ocial, \textbf{I}mage (both RGB and semantic) and \textbf{G}oal component. Our SAR model does not use any of these elements, as it relies only on the temporal component. We report a tick on the Image column only if a model uses image knowledge when predicting future time steps (\ie in the recurrent loop). This implies that \textsf{Y}-net has a tick while MG-GAN and Goal-SAR do not, as they only use the image in the goal module, leveraging strictly less information.
\vspace{-10pt}

\paragraph{inD dataset.}
Table~\ref{tab:short_term_ind} shows our results on the Intersection Drone dataset. We consider the evaluation protocol proposed in Bertugli \etal~\cite{Bertugli2021-acvrnn}. For a fair comparison, we train \textsf{Y}-net using the standard short-term evaluation protocol, as its authors only investigated this dataset in a long-term setting. The results are in line with the previous table, confirming the effectiveness of our architecture.

\paragraph{ETH/UCY dataset.}
Table~\ref{tab:eth_ucy} reports quantitative results for our methods on the traditional ETH/UCY benchmark. A similar trend with respect to the previous tables is outlined, with SAR performing on par with PECNet and outperforming both Goal-GAN and MG-GAN. It also outperforms by a fair margin Transformer-TF, which uses the standard encoder-decoder transformer architecture. This proves the effectiveness of our simpler network in modeling temporal sequences and confirms our intuition of using a recurrent self-attentive procedure. The addition of the goal module improves overall performance on average by $0.06$ and $0.22$ meters for ADE and FDE metrics, respectively. Goal-SAR performs on par with Trajectron++, AgentFormer and \textsf{Y}-net, reaching the second-best average ADE and FDE. We stress the fact that Goal-SAR uses image information only in the goal module, while Goal-GAN and \textsf{Y}-net also include it in the trajectory prediction step. These results prove the beneficial effects of encompassing goal information with a self-attentive mechanism, and that several approaches do not properly exploit the additional input image and social information when dealing with complex scenarios.

\vspace{0.5pt}

\subsection{Ablation Study}

Table~\ref{tab:ablation} reports an ablation study performed on SDD. We first investigate the effectiveness of our SAR model compared to other temporal backbones, obtained by replacing the temporal attention block with an LTSM and RNN cell, respectively. We then examine different possible concatenation mechanisms to combine the goal module and the temporal backbone. Skip-connection (\ie goal information concatenated both before and after the temporal attention module) outperforms both early and late fusion mechanisms. In all three cases, along with the estimated goal, we also concatenate additional information $\mathcal{I}_i$ (\ie last position, distance to the goal and current time step $t$). Our Goal-SAR architecture with skip-connection achieves the best overall performance.

\begin{figure}[th]
    \includegraphics[width=\columnwidth]{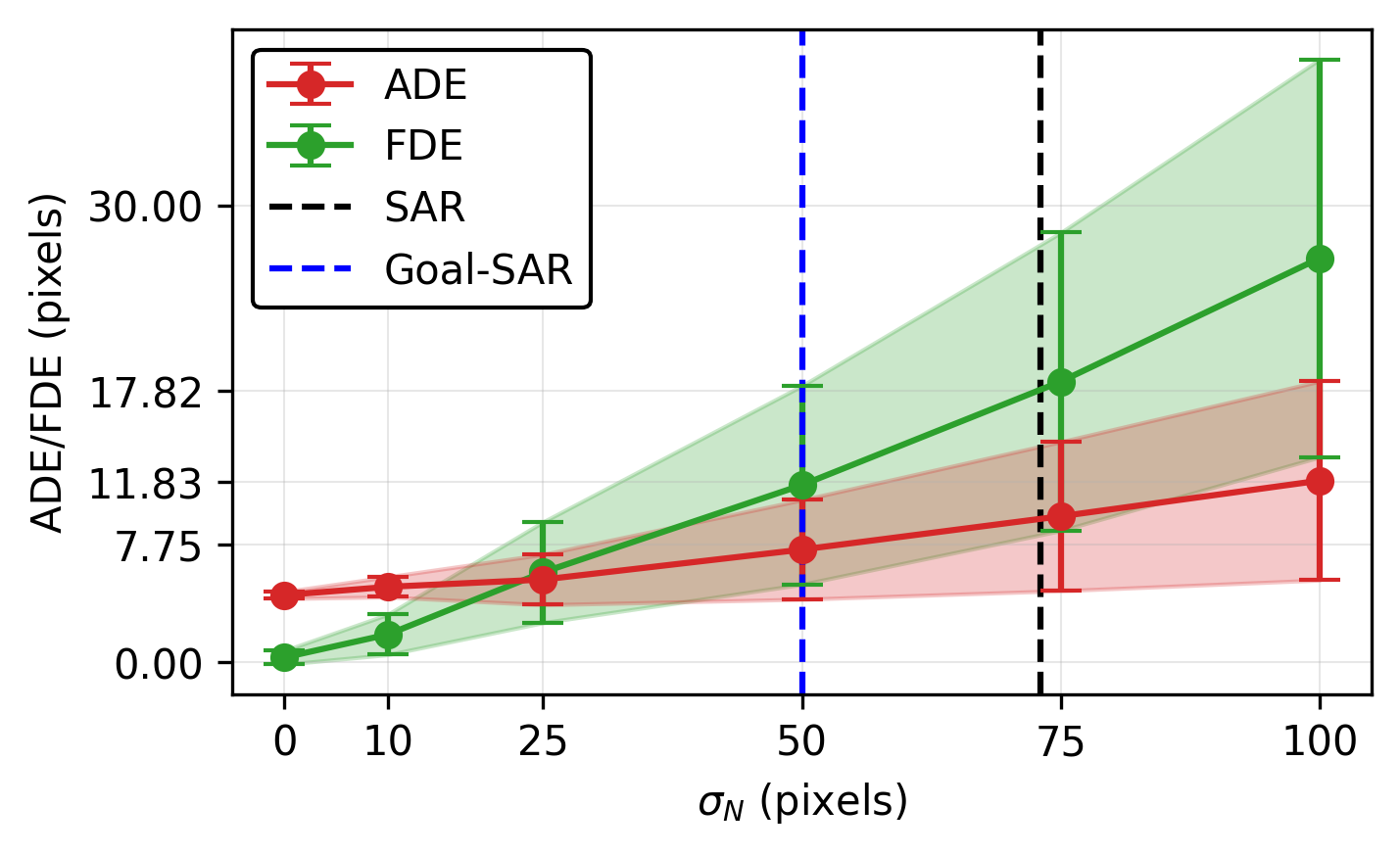}
    \caption{\textbf{Noisy ground truth goals.} ADE and FDE figures are obtained by feeding noisy ground truth destinations, with $\sigma_N$ representing noise amplitude. Results (in pixels) are obtained on SDD (error bars at $\pm 1$ standard deviation w.r.t all test trajectories) \vspace{-1pt}.
    }
    \label{fig:goal_gt_noise}
\end{figure}

\begin{figure}[ht]
    \includegraphics[width=\columnwidth]{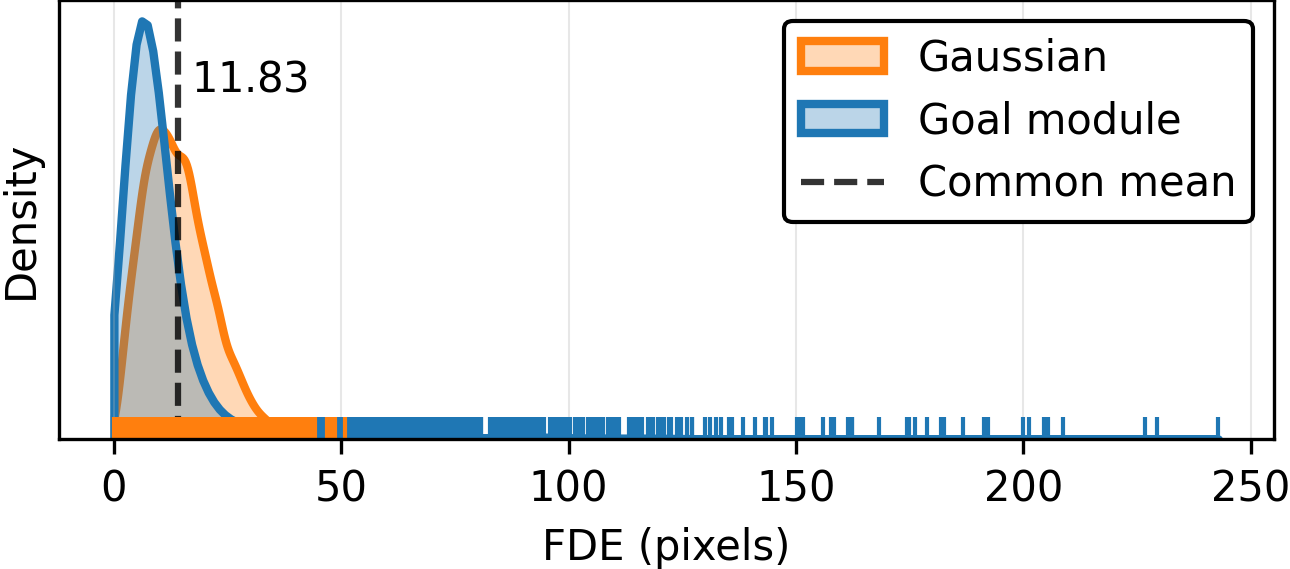}
    \caption{\textbf{Goal distribution.} We report the sampled distribution of the goal module outputs, and compare it to a Gaussian distribution with $\sigma_N = 50$ centered at the ground truth destinations. FDE metric (in pixels) is computed on SDD. \vspace{-5pt}}
    \label{fig:goal_fde}
\end{figure}

\begin{figure*}
\centering

    \subfloat[ETH]{\includegraphics[width=5cm, height=4.3cm]{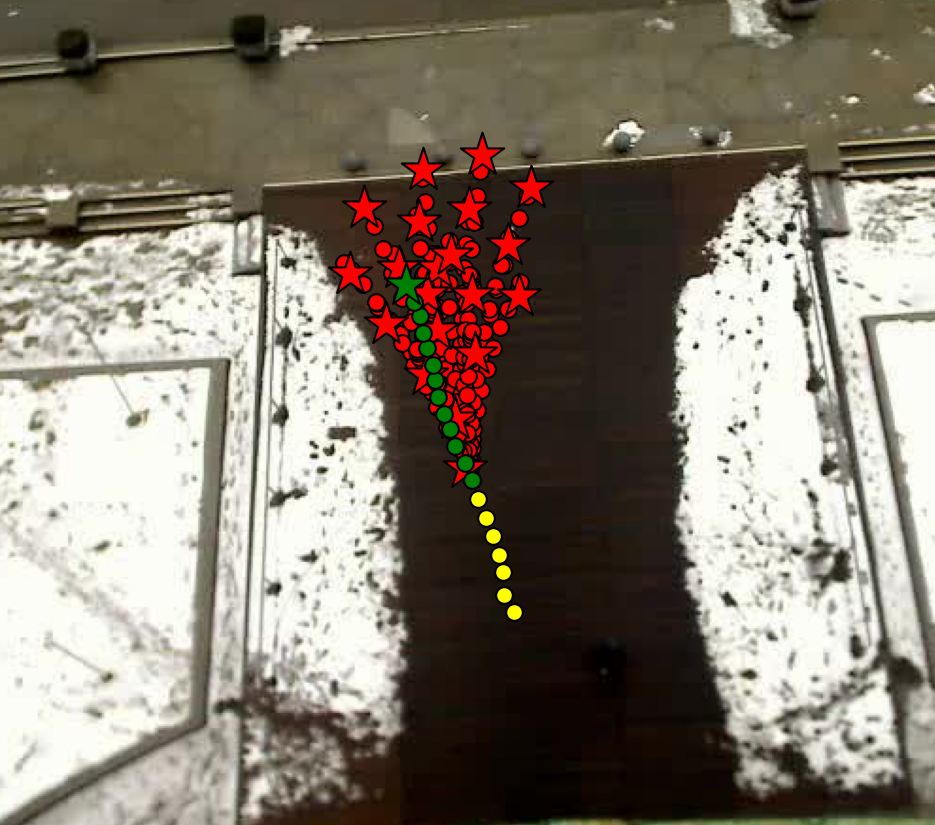}}
    \hspace{1em}
    \subfloat[Nexus]{\includegraphics[width=5cm, height=4.3cm]{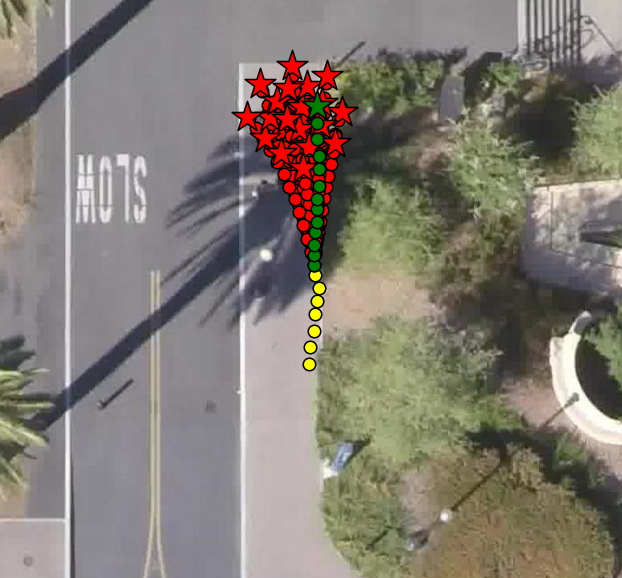}}
    \hspace{1em}
    \subfloat[inD]{\includegraphics[width=5cm, height=4.3cm]{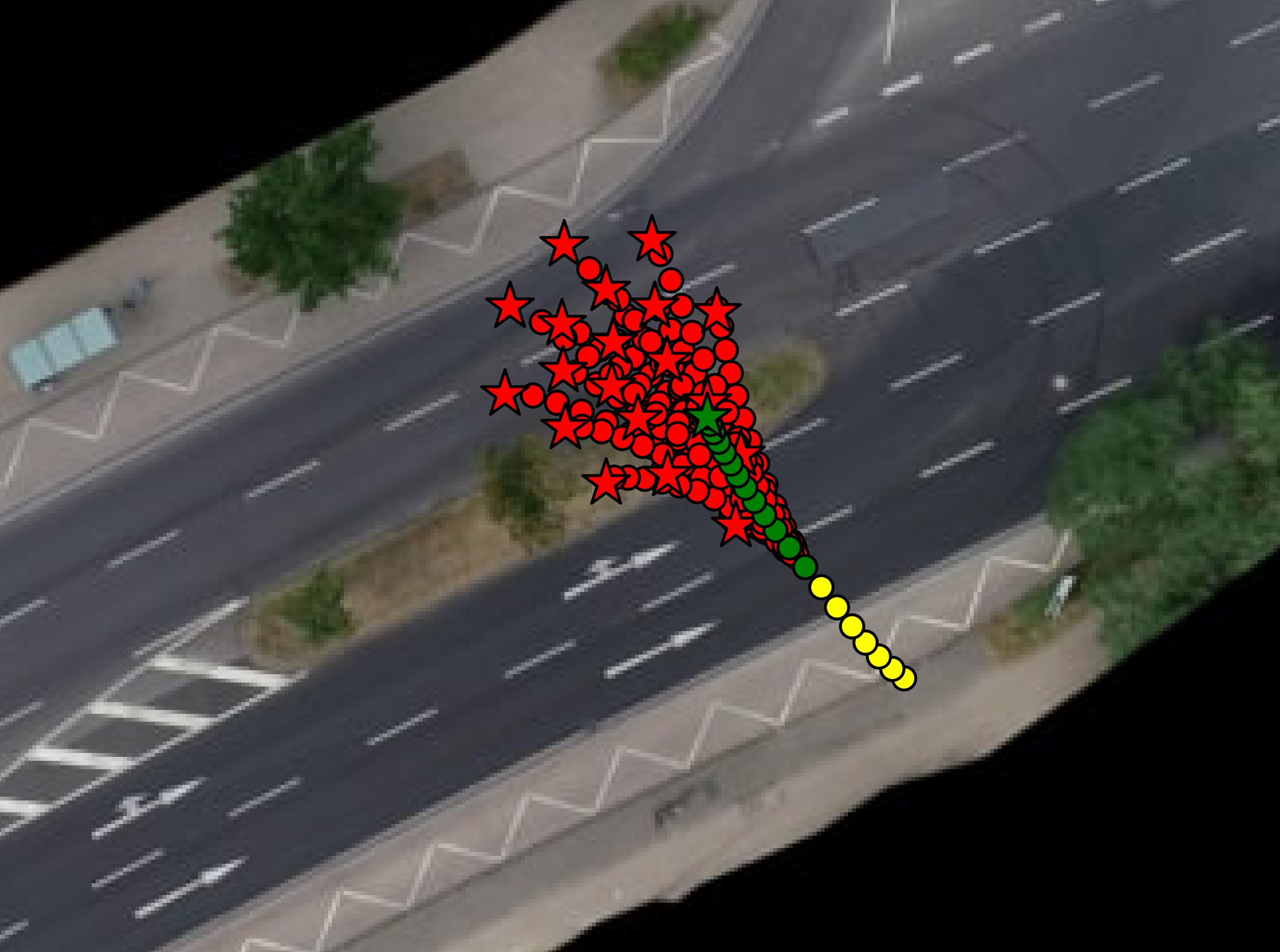}}
    \hspace{1em}
    \subfloat[Hyang]{\includegraphics[width=5cm, height=4.3cm]{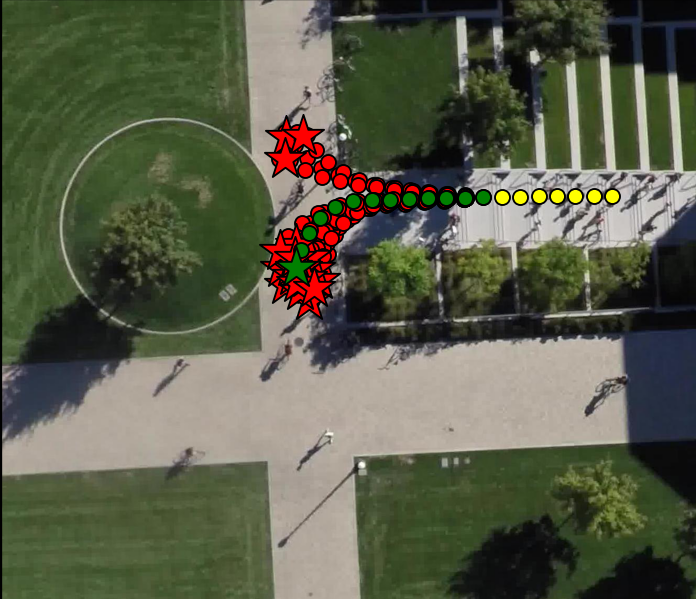}}
    \hspace{1em}
    \subfloat[ZARA1]{\includegraphics[width=5cm, height=4.3cm]{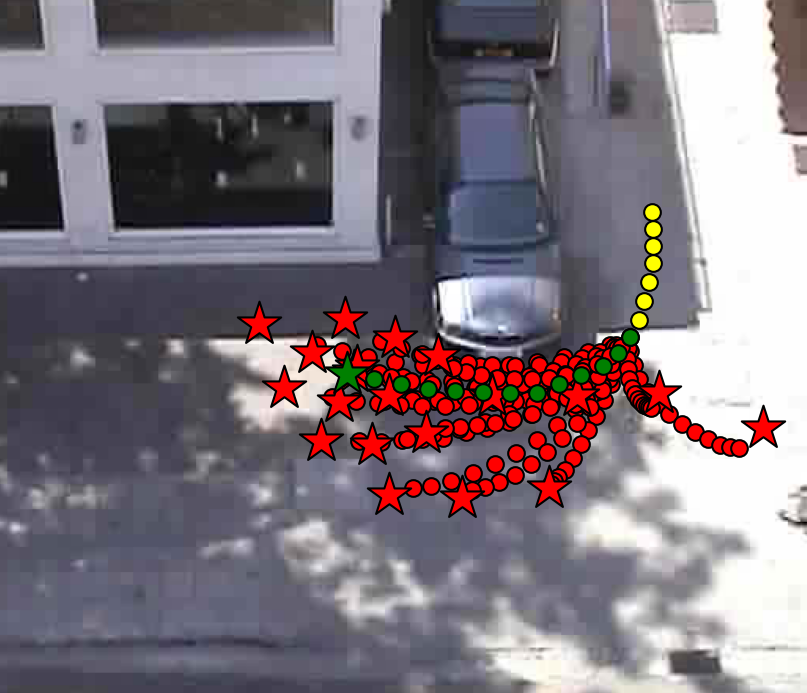}}
    \hspace{1em}
    \subfloat[UNIV]{\includegraphics[width=5cm, height=4.3cm]{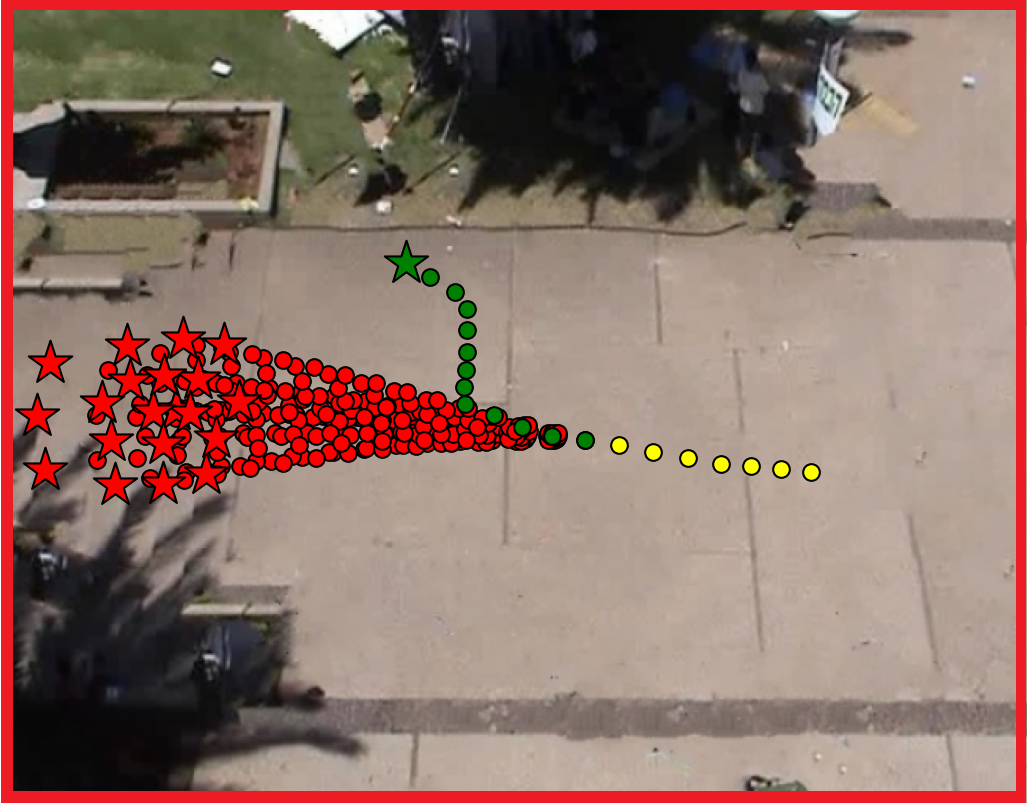}}

\caption{\textbf{Qualitative results.} Predicted trajectories (in red) and ground-truth locations (in green). Observed positions are depicted in yellow. The first row shows examples where a limited variability is predicted, while the second row highlights the multi-modal behavior induced by the goal module. The highlighted example (in red) shows a failure case.} 
\label{fig:qual_results}
\end{figure*}

To investigate the predictive power of the goal module and test the robustness of the overall architecture, we feed SAR with ground truth destinations instead of predicted ones, and then we gradually add random noise to ground truth goals.
We conduct these experiments on SDD. The results are reported in Fig.~\ref{fig:goal_gt_noise}, where $\sigma_{N}$ represents the standard deviation of a Gaussian distribution centered at the ground-truth goal. When $\sigma_{N} = 0$, \ie actual ground truth destinations are used, ADE metric is close to $4.42$ pixels while FDE metric is close to $0$.
By increasing the noise amplitude, performance deteriorates.
At $\sigma_{N} \approx 25$, ADE is overtaken by FDE, and at $\sigma_{N} \approx 50$, we obtain similar results to our goal module. It can be seen that our SAR backbone is comparable to a noisy ground-truth goal module with $\sigma_{N} \approx 73$. Beyond this value, goal information is no longer useful and its usage degrades prediction metrics.

We finally plot the goal module output distribution in FDE terms in Fig.~\ref{fig:goal_fde}, and compare it to an equivalent Gaussian distribution centered at the ground truth destinations ($\sigma_N = 50$), for each trajectory in the test set. It can be noted that, although the two distributions have the same mean (\ie FDE $=11.83$), the goal module predictions are more scattered. Indeed, while the majority of the goal distribution is shifted towards zero error, there exist some outliers, and a few of them are up to $200+$ pixels away from the real goals. This observation suggests that the goal module has a margin of improvement with respect to some edge cases.

\subsection{Qualitative Results}
Fig.~\ref{fig:qual_results} shows some qualitative results obtained with Goal-SAR. Fig.~\ref{fig:qual_results} (a)-(c) shows a few examples of easily predictable trajectories on different datasets. In (d) and (e) multi-modality is well captured by the goal module, as these paths appear likely to occur in the future. Finally, in Fig.~\ref{fig:qual_results} (f), we report a failure case in which, due to a sudden direction change, our model fails to predict the future positions.

\subsection{Limitations}
\label{sec:limitations}
Quantitative and qualitative results demonstrate that our method is able to efficiently use goal information in a recurrent architecture. This information helps our recurrent backbone to produce scene-compliant predictions, that would otherwise be completely unaware of the surrounding context. We deliberately decided not to consider social interactions in our analysis to keep our architecture as straightforward as possible. Nevertheless, we hypothesize that it is possible to improve our results even further if both human-human relations and scene semantics are included.


\section{Conclusion}
\label{sec:conclusion}
To predict future human positions, we propose a simple yet effective attention-based recurrent architecture that processes temporal dependencies. When combined with a scene-aware goal-estimation module, our model is able to estimate likely scene-complaint trajectories with an increased multi-modality capability. We demonstrate its efficacy in reducing standard error metrics in a stochastic setting, when two or more paths are plausible. Finally, we emphasize the fact that our architecture is fairly uncomplicated, does not leverage social information, and scene knowledge is only processed in the goal module. Despite using less information, the proposed approach performs on par with many state-of-the-art models, and even shows slight improvements in some scenarios.

\paragraph*{Acknowledgements.} 
This work was supported in part by the PRIN-17 PREVUE project from the Italian MUR (CUP E94I19000650001).
We also acknowledge the HPC resources of UniPD -- DM and CAPRI clusters -- and NVIDIA for their donation of GPUs used in this research.

{\small

\begin{thebibliography}{10}\itemsep=-1pt

\bibitem{Social-lstm}
Alexandre Alahi, Kratarth Goel, Vignesh Ramanathan, Alexandre Robicquet, Li
  Fei-Fei, and Silvio Savarese.
\newblock {Social LSTM}: Human trajectory prediction in crowded spaces.
\newblock In {\em CVPR}, 2016.

\bibitem{albrecht2021interpretable}
Stefano~V. Albrecht, Cillian Brewitt, John Wilhelm, Balint Gyevnar, Francisco
  Eiras, Mihai Dobre, and Subramanian Ramamoorthy.
\newblock Interpretable goal-based prediction and planning for autonomous
  driving.
\newblock In {\em ICRA}, 2021.

\bibitem{KTMP16}
Lamberto Ballan, Francesco Castaldo, Alexandre Alahi, Francesco Palmieri, and
  Silvio Savarese.
\newblock Knowledge transfer for scene-specific motion prediction.
\newblock In {\em ECCV}, 2016.

\bibitem{Bertugli2021-acvrnn}
Alessia Bertugli, Simone Calderara, Pasquale Coscia, Lamberto Ballan, and Rita
  Cucchiara.
\newblock {AC-VRNN: Attentive Conditional-VRNN} for multi-future trajectory
  prediction.
\newblock {\em Computer Vision and Image Understanding}, 2021.

\bibitem{log_sum_trick}
Pierre Blanchard, Desmond~J Higham, and Nicholas~J Higham.
\newblock {Accurately computing the log-sum-exp and softmax functions}.
\newblock {\em IMA Journal of Numerical Analysis}, 2020.

\bibitem{inDdataset}
Julian Bock, Robert Krajewski, Tobias Moers, Steffen Runde, Lennart Vater, and
  Lutz Eckstein.
\newblock The ind dataset: A drone dataset of naturalistic road user
  trajectories at german intersections.
\newblock {\em IEEE Intelligent Vehicles Symposium}, 2020.

\bibitem{Choi_2019_ICCV}
Chiho Choi and Behzad Dariush.
\newblock Looking to relations for future trajectory forecast.
\newblock In {\em ICCV}, 2019.

\bibitem{IVC18}
Pasquale Coscia, Francesco Castaldo, Francesco Palmieri, Alexandre Alahi,
  Silvio Savarese, and Lamberto Ballan.
\newblock Long-term path prediction in urban scenarios using circular
  distributions.
\newblock {\em Image and Vision Computing}, 2018.

\bibitem{Dendorfer_2021_ICCV}
Patrick Dendorfer, Sven Elflein, and Laura Leal-Taix\'e.
\newblock {MG-GAN}: A multi-generator model preventing out-of-distribution
  samples in pedestrian trajectory prediction.
\newblock In {\em ICCV}, 2021.

\bibitem{dendorfer2020goalgan}
Patrick Dendorfer, Aljoša Ošep, and Laura Leal-Taixé.
\newblock {Goal-GAN}: Multimodal trajectory prediction based on goal position
  estimation.
\newblock In {\em ACCV}, 2020.

\bibitem{goal_rl2}
Carlos Florensa, David Held, Xinyang Geng, and Pieter Abbeel.
\newblock Automatic goal generation for reinforcement learning agents.
\newblock In {\em ICML}, 2018.

\bibitem{goal_rl}
Dibya Ghosh, Abhishek Gupta, and Sergey Levine.
\newblock Learning actionable representations with goal-conditioned policies.
\newblock In {\em ICLR}, 2019.

\bibitem{giuliari2020transformer}
Francesco Giuliari, Irtiza Hasan, Marco Cristani, and Fabio Galasso.
\newblock Transformer networks for trajectory forecasting.
\newblock In {\em ICPR}, 2020.

\bibitem{social-gan}
Agrim Gupta, Justin Johnson, Li Fei-Fei, Silvio Savarese, and Alexandre Alahi.
\newblock {Social GAN}: Socially acceptable trajectories with generative
  adversarial networks.
\newblock In {\em CVPR}, 2018.

\bibitem{social_force}
Dirk Helbing and P\'eter Moln\'ar.
\newblock Social force model for pedestrian dynamics.
\newblock {\em Phys. Rev. E}, 1995.

\bibitem{Huang_2019_ICCV}
Yingfan Huang, Huikun Bi, Zhaoxin Li, Tianlu Mao, and Zhaoqi Wang.
\newblock Stgat: Modeling spatial-temporal interactions for human trajectory
  prediction.
\newblock In {\em ICCV}, 2019.

\bibitem{DBLP:conf/icml/Kaelbling93}
Leslie~Pack Kaelbling.
\newblock Hierarchical learning in stochastic domains: Preliminary results.
\newblock In {\em ICML}, 1993.

\bibitem{kipf2017semi}
Thomas~N. Kipf and Max Welling.
\newblock Semi-supervised classification with graph convolutional networks.
\newblock In {\em ICLR}, 2017.

\bibitem{Kitani_2012_ECCV}
Kris~M. Kitani, Brian~D. Ziebart, James~Andrew Bagnell, and Martial Hebert.
\newblock Activity forecasting.
\newblock In {\em ECCV}, 2012.

\bibitem{trajnetpp}
Parth Kothari, Sven Kreiss, and Alexandre Alahi.
\newblock Human trajectory forecasting in crowds: A deep learning perspective.
\newblock {\em IEEE Transactions on Intelligent Transportation Systems}, 2021.

\bibitem{lerner}
Alon Lerner, Yiorgos Chrysanthou, and Dani Lischinski.
\newblock Crowds by example.
\newblock {\em Computer Graphics Forum}, 2007.

\bibitem{Liang_2020_CVPR}
Junwei Liang, Lu Jiang, Kevin Murphy, Ting Yu, and Alexander Hauptmann.
\newblock The garden of forking paths: Towards multi-future trajectory
  prediction.
\newblock In {\em CVPR}, 2020.

\bibitem{Lisotto_2019_ICCV}
Matteo Lisotto, Pasquale Coscia, and Lamberto Ballan.
\newblock Social and scene-aware trajectory prediction in crowded spaces.
\newblock In {\em ICCVW}, 2019.

\bibitem{liu2020snce}
Yuejiang Liu, Qi Yan, and Alexandre Alahi.
\newblock Social nce: Contrastive learning of socially-aware motion
  representations.
\newblock In {\em ICCV}, 2021.

\bibitem{Mangalam_2021_ICCV}
Karttikeya Mangalam, Yang An, Harshayu Girase, and Jitendra Malik.
\newblock From goals, waypoints \& paths to long term human trajectory
  forecasting.
\newblock In {\em ICCV}, 2021.

\bibitem{mangalam2020pecnet}
Karttikeya Mangalam, Harshayu Girase, Shreyas Agarwal, Kuan-Hui Lee, Ehsan
  Adeli, Jitendra Malik, and Adrien Gaidon.
\newblock It is not the journey but the destination: Endpoint conditioned
  trajectory prediction.
\newblock In {\em ECCV}, 2020.

\bibitem{mantra}
Francesco Marchetti, Federico Becattini, Lorenzo Seidenari, and Alberto
  Del~Bimbo.
\newblock {MANTRA}: Memory augmented networks for multiple trajectory
  prediction.
\newblock In {\em CVPR}, 2020.

\bibitem{Mohamed_2020_CVPR}
Abduallah Mohamed, Kun Qian, Mohamed Elhoseiny, and Christian Claudel.
\newblock {Social-STGCNN}: A social spatio-temporal graph convolutional neural
  network for human trajectory prediction.
\newblock In {\em CVPR}, 2020.

\bibitem{pellegrini}
Stefano Pellegrini, Andreas Ess, Konrad Schindler, and Luc Van~Gool.
\newblock You'll never walk alone: Modeling social behavior for multi-target
  tracking.
\newblock In {\em ICCV}, 2009.

\bibitem{sdd}
Alexandre Robicquet, Amir Sadeghian, Alexandre Alahi, and Silvio Savarese.
\newblock Learning social etiquette: Human trajectory understanding in crowded
  scenes.
\newblock In {\em ECCV}, 2016.

\bibitem{u-net}
Olaf Ronneberger, Philipp Fischer, and Thomas Brox.
\newblock {U-Net}: Convolutional networks for biomedical image segmentation.
\newblock In {\em MICCAI}, 2015.

\bibitem{rudenko_survey}
Andrey Rudenko, Luigi Palmieri, Michael Herman, Kris~M. Kitani, Dariu~M.
  Gavrila, and Kai~O. Arras.
\newblock Human motion trajectory prediction: a survey.
\newblock {\em IJRR}, 2020.

\bibitem{Sadeghian_2019_CVPR}
Amir Sadeghian, Vineet Kosaraju, Ali Sadeghian, Noriaki Hirose, Hamid
  Rezatofighi, and Silvio Savarese.
\newblock Sophie: An attentive {GAN} for predicting paths compliant to social
  and physical constraints.
\newblock In {\em CVPR}, 2019.

\bibitem{Sadeghian_2018_ECCV}
Amir Sadeghian, Ferdinand Legros, Maxime Voisin, Ricky Vesel, Alexandre Alahi,
  and Silvio Savarese.
\newblock {CAR-Net}: Clairvoyant attentive recurrent network.
\newblock In {\em ECCV}, 2018.

\bibitem{trajectronpp}
Tim Salzmann, Boris Ivanovic, Punarjay Chakravarty, and Marco Pavone.
\newblock Trajectron++: Multi-agent generative trajectory forecasting with
  heterogeneous data for control.
\newblock In {\em ECCV}, 2020.

\bibitem{constant_velocity}
Christoph Schöller, Vincent Aravantinos, Florian Lay, and Alois Knoll.
\newblock What the constant velocity model can teach us about pedestrian motion
  prediction.
\newblock {\em Robotics and Automation Letters}, 2020.

\bibitem{vaswani_attention}
Ashish Vaswani, Noam Shazeer, Niki Parmar, Jakob Uszkoreit, Llion Jones,
  Aidan~N. Gomez, \L{}ukasz Kaiser, and Illia Polosukhin.
\newblock Attention is all you need.
\newblock In {\em NeurIPS}, 2017.

\bibitem{Vemula2017SocialAM}
Anirudh Vemula, Katharina Muelling, and Jean Oh.
\newblock Social attention: Modeling attention in human crowds.
\newblock In {\em ICRA}, 2017.

\bibitem{9046288}
Zonghan Wu, Shirui Pan, Fengwen Chen, Guodong Long, Chengqi Zhang, and
  Philip~S. Yu.
\newblock A comprehensive survey on graph neural networks.
\newblock {\em IEEE Transactions on Neural Networks and Learning Systems},
  2021.

\bibitem{ss-lstm}
Hao Xue, Du~Q. Huynh, and Mark Reynolds.
\newblock {SS-LSTM}: A hierarchical {LSTM} model for pedestrian trajectory
  prediction.
\newblock In {\em WACV}, 2018.

\bibitem{NEURIPS2019_dc6a7e65}
Zhilin Yang, Zihang Dai, Yiming Yang, Jaime Carbonell, Russ~R Salakhutdinov,
  and Quoc~V Le.
\newblock {XLNet}: Generalized autoregressive pretraining for language
  understanding.
\newblock In {\em NeurIPS}, 2019.

\bibitem{YuMa2020Spatio}
Cunjun Yu, Xiao Ma, Jiawei Ren, Haiyu Zhao, and Shuai Yi.
\newblock Spatio-temporal graph transformer networks for pedestrian trajectory
  prediction.
\newblock In {\em ECCV}, 2020.

\bibitem{Yuan_2021_ICCV}
Ye Yuan, Xinshuo Weng, Yanglan Ou, and Kris~M. Kitani.
\newblock Agentformer: Agent-aware transformers for socio-temporal multi-agent
  forecasting.
\newblock In {\em ICCV}, 2021.

\bibitem{zhang2019srlstm}
Pu Zhang, Wanli Ouyang, Pengfei Zhang, Jianru Xue, and Nanning Zheng.
\newblock {SR-LSTM}: State refinement for {LSTM} towards pedestrian trajectory
  prediction.
\newblock In {\em CVPR}, 2019.

\bibitem{zhao2020tnt}
Hang Zhao, Jiyang Gao, Tian Lan, Chen Sun, Benjamin Sapp, Balakrishnan
  Varadarajan, Yue Shen, Yi Shen, Yuning Chai, Cordelia Schmid, Congcong Li,
  and Dragomir Anguelov.
\newblock {TNT: Target-driven trajectory prediction}.
\newblock In {\em CoRL}, 2020.

\bibitem{Zhao_2021_ICCV}
He Zhao and Richard~P. Wildes.
\newblock Where are you heading? dynamic trajectory prediction with expert goal
  examples.
\newblock In {\em ICCV}, 2021.

\bibitem{ziebart2009planning}
Brian~D. Ziebart, Nathan Ratliff, Garratt Gallagher, Christoph Mertz, Kevin
  Peterson, J~Andrew Bagnell, Martial Hebert, Anind~K Dey, and Siddhartha
  Srinivasa.
\newblock Planning-based prediction for pedestrians.
\newblock In {\em IROS}, 2009.

\end{thebibliography}

}

\end{document}